\documentclass[journal]{IEEEtran}
\usepackage{cite}

  \usepackage[pdftex]{graphicx}
\usepackage[cmex10]{amsmath}
\usepackage{amsfonts}
\usepackage{amssymb}
\usepackage{algorithm}
\usepackage{algorithmic}
  \usepackage[caption=false,font=scriptsize]{subfig}
\usepackage{fixltx2e}

\begin{document}
%
\title{A Unified Tensor-based Active Appearance Face Model}
%
%
%

\author{Zhen-Hua~Feng,
~\IEEEmembership{Member,~IEEE,}
        Josef~Kittler,
~\IEEEmembership{Life~Member,~IEEE,}
	William~Christmas,
	and~Xiao-Jun~Wu,
~\IEEEmembership{Member,~IEEE}
}

\newcommand{\eg}{\textit{e.g.~}}
\newcommand{\ie}{\textit{i.e.~}}
\newcommand{\etal}{\textit{et. al.~}}
\maketitle

\begin{abstract}
Appearance variations result in many difficulties in face image analysis. To deal with this challenge, we present a Unified Tensor-based Active Appearance Model (UT-AAM) for jointly modelling the geometry and texture information of 2D faces. For each type of face information, namely shape and texture, we construct a unified tensor model capturing all relevant appearance variations. This contrasts with the variation-specific models of the classical tensor AAM. To achieve the unification across pose variations, a strategy for dealing with self-occluded faces is proposed to obtain consistent shape and texture representations of pose-varied faces. In addition, our UT-AAM is capable of constructing the model from an incomplete training dataset, using tensor completion methods. Last, we use an effective cascaded-regression-based method for UT-AAM fitting. With these advancements, the utility of UT-AAM in practice is considerably enhanced. As an example, we demonstrate the improvements in training facial landmark detectors through the use of UT-AAM to synthesise a large number of virtual samples. Experimental results obtained using the Multi-PIE and 300-W face datasets demonstrate the merits of the proposed approach.
\end{abstract}


%
\IEEEpeerreviewmaketitle

\section{Introduction}
Geometry and texture information of an object plays important roles in a variety of computer vision and pattern recognition tasks.
For face image analysis, face geometry and texture provide important clues for processing and interpreting human faces, \eg face recognition, emotion analysis as well as face animation.
In general, the face geometry (or face shape) is in the form of a vector consisting of the 2D coordinates of a set of pre-defined key points with semantic meaning, \eg nose tip, eye corners and face outline; 
and the texture refers to pixel intensities of the face.
In automatic face analysis systems, face shape and texture are usually used jointly. 
Face shape is crucial for aligning/registering face texture as a prerequisite to extracting meaningful textural features for the following analysis steps in the processing pipeline~\cite{liao2013partial,taigman2014deepface,masi2016pose}.

\begin{figure}[!t]
	\centering
	\includegraphics[trim = 0mm 55mm 80mm 0mm, clip,width=.9\linewidth]{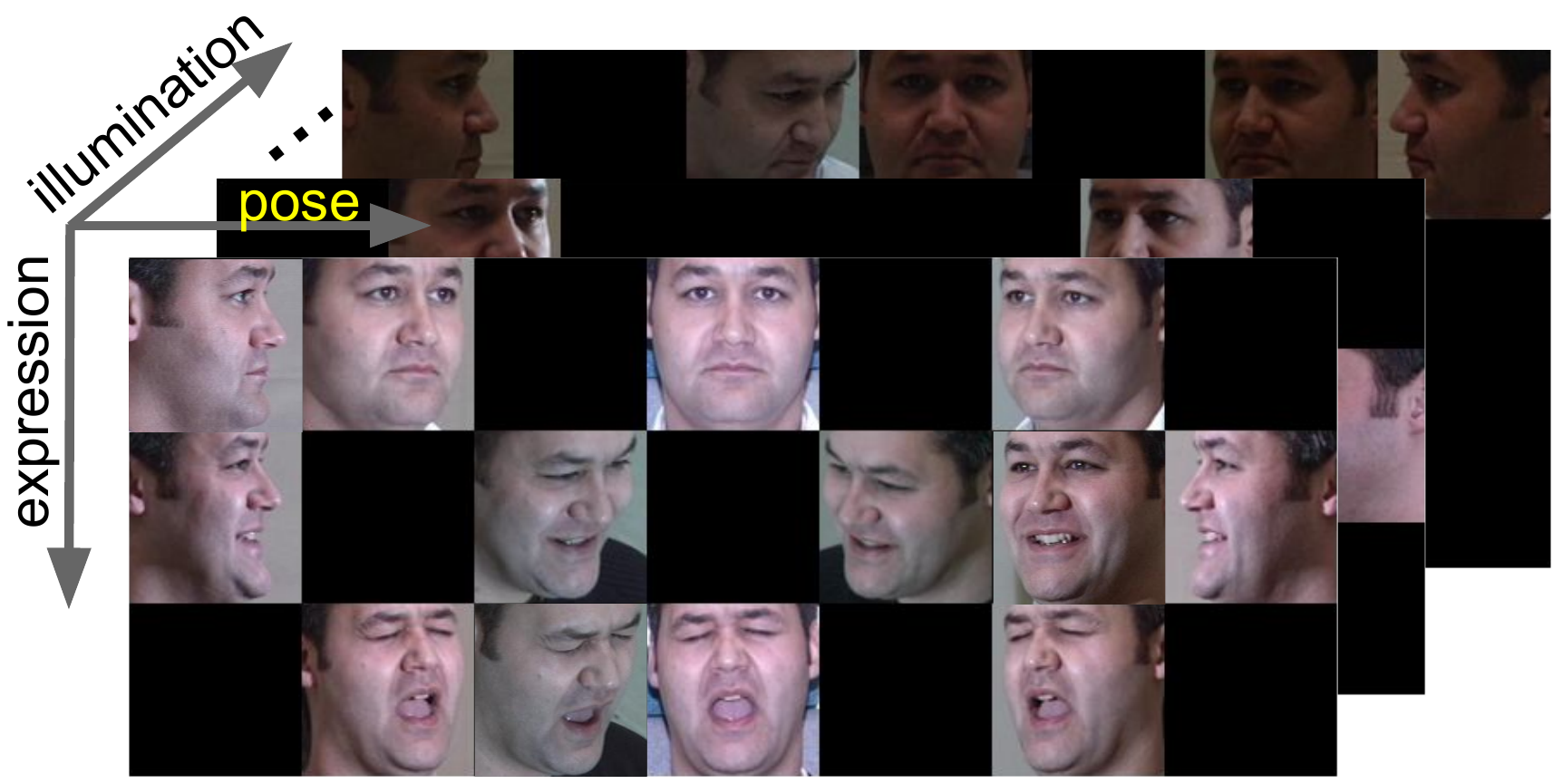}
	\caption{Example images of a subject in an incomplete training dataset with pose, expression and illumination variations.}
	\label{fig_1}
\end{figure}

A well-known approach that is capable of recovering and representing shape and texture information of faces is the Active Appearance Model~(AAM)~\cite{Cootes1998, Cootes2001}, which has also been widely used for many other applications such as medical image analysis.
However, to build such a face model is non-trivial, due to a wide range of appearance variations in pose, expression, illumination and occlusion~\cite{Gonzalez-Mora2007,Lee2009}.
One challenge is how to represent the geometry and texture information of a face compactly.
To this end, tensor-based AAM (T-AAM)~\cite{Lee2008,Lee2009} has been proposed.
T-AAM decomposes the original shape and texture space into a set of factor-related subspaces using Higher-Order Single Value Decomposition (HOSVD).
Its success derives from the capability of multilinear subspace analysis to decouple multi-factor signals.
However, the use of T-AAM has some issues in practice:
1) To build a tensor-based model, we have to collect a large number of training samples with different variations. 
For example, to build a T-AAM consisting of 10 pose, 10 expression and 10 illumination variations, the required number of training images is $10^3$ per subject.
It is difficult to capture and manually annotate so many face images.
As shown in Fig.~\ref{fig_1}, we may have missing training samples in a dataset.
2) To fit a new face image, T-AAM generates a number of variation-specific models. 
Then, a tensor-based state estimation method is used to select a suitable variation-specific model for T-AAM fitting.
This task is appearance-based and relies on a face/eye detector; hence the accuracy of state estimation cannot be guaranteed.
3) T-AAM does not consider the problem of the key points located in the self-occluded facial parts that are invisible in a 2D face image, which is usually caused by extreme pose variations.
T-AAM has avoided this problem by focusing on relatively slight pose deviations (up to $22.5^\circ$) from the frontal one.
4) The classical T-AAM fitting algorithm is gradient-descent-based~\cite{Lee2009, Matthews2004}, and depends on the estimation of the state of each variation mode for a new face.
It can easily be trapped in local minima, especially when the state estimation of a face is inaccurate.

To address these issues, we present a Unified Tensor-based AAM (UT-AAM). In contrast to the classical T-AAM, our UT-AAM has four main innovations:
\begin{itemize}
\item The unification is achieved by adopting two measures.
First, we create a single model across different variation modes, rather than using a number of variation-specific models.
Second, we tackle the problem of self-occlusion of faces from large-scale pose variations by proposing a uniform landmarking strategy.

\item Enabling the proposed UT-AAM to be constructed from an incomplete dataset with missing training samples.
To this end, we investigate tensor completion algorithms to construct our UT-AAM from incomplete training datasets.
Furthermore, a new initialisation method is developed to obtain a better reconstruction of missing training samples.

\item We develop a learning-based model fitting algorithm.
The proposed algorithm does not require prior knowledge of the state of each variation mode for fitting a new face, hence we do not have to perform state estimation before UT-AAM fitting.
Moreover, according to our experimental results obtained from the Multi-PIE dataset, the proposed algorithm offers more accurate fitting results.

\item We demonstrate the merit of the proposed UT-AAM in synthesising realistic virtual faces with arbitrary pose variations.
We perform data augmentation for the training of a facial landmark detector, using a large number of synthetic faces.
Experimental results obtained on the 300-W dataset validate the advancements of the use of virtual faces synthesised by our UT-AAM.
\end{itemize}

The rest of this paper is organised as follows: 
Section~\ref{Sec_RelatedWork} gives a brief introduction to related work.
Section~\ref{Sec_TAAM} overviews the classical AAM and T-AAM algorithms. 
The proposed UT-AAM algorithm is presented in Section~\ref{Sec_UTAAM},
and the experimental results are reported in Section~\ref{Sec_Evaluation}. 
Last, some conclusions are presented in Section~\ref{Sec_Conclusion}.

\section{Related work}
\label{Sec_RelatedWork}
To obtain the geometry and texture information of 2D faces, a variety of methods have been developed during the past decades, \eg Active Shape Model (ASM)~\cite{Cootes1995}, Active Appearance Model (AAM)~\cite{Cootes1998,Cootes2001}, Constrained Local Model (CLM)~\cite{cristinacce2006feature,cristinacce2008automatic} and Cascaded Regression (CR-) based facial landmark detection methods~\cite{dollar2010cascaded,xiong2013supervised,cao2014face,feng2017dynamic,feng2017cvprw}.
Among these algorithms, AAM is capable of jointly modelling the shape and texture information of faces.
ASM, CLM and CR-based approaches are mainly used for obtaining the shape information conveyed by facial landmarks. 
However, fitting AAM to 2D faces is non-trivial, especially for faces exhibiting a wide range of appearance variations.
The developments of AAM addressing this issue can be divided into two categories.
In the \textit{first} category, the aim is to improve the structure of the underlying AAM models for better representation of shape or texture information, which is also expected to benefit the subsequent model fitting phase.
The \textit{second} category focuses on developing fitting algorithms that generalise well for unseen faces with higher accuracy and less computational cost.

\subsection{Model structure}
A generative AAM has two PCA-based models for shape and texture, respectively.
AAM is capable of generating arbitrary face instances by adjusting model parameters.
The texture model of an AAM is usually constructed from raw pixel intensities.
Hence an AAM fitting algorithm based on the optimisation of a cost function using texture information is sensitive to appearance variations.
To eliminate this dependence, the use of variation-invariant features has been suggested, such as multi-band Value-Hue-Edge~\cite{stegmann2003multi}, image gradients~\cite{kittipanya2006effect}, Haar-Like features~\cite{liu2009discriminative} and image gradient orientations~\cite{tzimiropoulos2014active,antonakos2015feature}.
In addition, local features have been successfully used to fit a shape model to faces, such as local profiles in ASM and local patches in CLM.
More recently, variation-invariant local descriptors have been very popular in CR-based facial landmark detection.
To extract local features, we usually apply a local descriptor, such as HOG~\cite{yan2013learn,huber2015fitting,feng2015cascaded}, SIFT~\cite{xiong2013supervised}, local pixel difference~\cite{cao2014face,ren2014face} or Sparse Auto-Encoder~\cite{feng2015random, zhang2014coarse}, in the neighbourhood of facial landmarks.

Another way to modify the model structure is to rely on different underlying methods.
The representation capacity of a PCA-based AAM, constructed from a small number of training samples, is limited.
For unseen faces, the PCA-based model may miss some details and in consequence it is not able to represent complex faces faithfully.
To cope with this issue, more advanced techniques, \eg kernel methods~\cite{Romdhani1999,Romdhani2000,hamsici2009active} and Deep Boltzmann Machines~\cite{nhan2015beyond,duong2016deep}, have been suggested for model construction.
Note that this limitation can also be addressed by using more PCA components trained from more samples with a wide range of appearance variations, as demonstrated by~\cite{tzimiropoulos2013optimization,booth20163d}.
Besides the representation capacity of AAM, a more important issue is how to construct a compact and structural model.
A common way to do this is to use multi-view models, \eg the View-based AAM (V-AAM)~\cite{Cootes2000,Romdhani2000}.
However, this strategy is resource- and time-consuming because we have to construct, store and fit multiple models to a face.
As an alternative, Bilinear AAM (B-AAM) constructs a unified model across pose variations~\cite{Gonzalez-Mora2007}.
Nevertheless, both V-AAM and B-AAM can only deal with a single variation type among pose, expression and illumination modes.

To analyse multi-factor signals, in recent years, tensor algebra has produced impressive results for  many computer vision and pattern recognition applications~\cite{Vasilescu2002, vlasic2005face, he2006tensor, vasilescu2007multilinear, Qi_2016_CVPR}.
As far as human faces are concerned, an instantaneous observation of the appearance of a subject depends on many factors (as shown in Fig.~\ref{fig_1}), hence its representation is naturally amended to a tensor-based modelling. 
Tensors have been shown to be a powerful tool for overcoming difficulties posed by appearance variations in AAM modelling and fitting.
For instance, T-AAM builds a set of variation-specific AAMs and then fit the corresponding model to an input image by estimating the states of pose, illumination and expression of the face~\cite{Lee2008,Lee2009}.
It should be noted that the classical AAM, V-AAM and B-AAM can be viewed as degenerations of T-AAM.
However, as discussed in the last section, the use of T-AAM has some issues in practice.
To tackle these issues, we propose a new framework that introduces a unified tensor-based model across different variation modes.

\subsection{Model fitting}
The target of AAM fitting is to find the parameters of a face model which best reconstruct the shape and texture information of a new face image.
This is a non-linear multi-variable optimisation problem.
Depending on the underlying model, the fitting algorithms can either be gradient-descent-based or learning-based.
In a gradient-descent-based method, a cost function, designed to recover the model parameters, is optimised by calculating its partial derivatives with respect to the model parameters.
For example, in classical AAM, the cost is the pixel intensity difference between the model and an input image.
Given their initial values, the model parameters are iteratively updated using gradient descent, such as steepest descent and Gauss-Newton methods~\cite{sclaroff1998active,Sung2007,Matthews2004}.
Although some techniques can be used to speed up the fitting phase and prevent the algorithms falling into local minima (\eg a pyramid scheme~\cite{Cootes2001}), the iterative computation of partial derivatives is time-consuming.
A distinctive milestone in the history of development is the inverse compositional AAM fitting algorithm that neatly avoids the iterative computation of partial derivatives by inversely compositing model parameters in a projecting-out image space~\cite{Matthews2004, matthews2004template}. 
This seminal work has greatly improved the speed of AAM fitting and consequently broadened its applications.
The extended approaches based on the inverse compositional fitting algorithm have achieved competitive results even for faces in the wild~\cite{tzimiropoulos2013optimization, antonakos2015feature, alabort2016unified}.

An alternative way of AAM fitting is to use machine learning techniques, either classification-based or regression-based.
For a given observation, the goal of classification-based methods is to maximise the probability of model parameters.
For example, Liu considered AAM fitting as a binary classification problem and achieved favourable fitting results with Haar-Like features~\cite{liu2009discriminative}.
As another example, the Support Vector Machine (SVM) has been used as a local expert to identify the best candidate point in the vicinity of a face key point in a CLM-based framework~\cite{wang2008enforcing,lucey2009efficient}.
Unlike classification-based methods, regression-based approaches estimate the partial derivatives by learning from a set of examples. 
For instance, Cootes \textit{et al.} reported great success with linear regression in their early studies of AAM~\cite{Cootes1998}, in which they assumed that there is a constant linear relationship between fitting residuals and parameters updates.
However, the simple linear regressor is incapable of solving such a complicated non-linear multi-variable optimisation problem~\cite{Matthews2004}.
To deal with this issue, more powerful regression methods have been used, such as Canonical Correlation Analysis (CCA)~\cite{Donner2006}, decision stumps~\cite{saragih2007nonlinear} and random forests~\cite{sauer2011accurate,cootes2012robust}.
More recently, regression-based approaches, in particular the cascaded regression, have been widely used for shape model fitting (also known as facial landmark detection), delivering promising results in both constrained and unconstrained scenarios.
The key idea of cascaded regression is to form a strong regressor by cascading a set of weak regressors in series.
A weak regressor in cascaded regression could be any regression method, such as linear regression~\cite{xiong2013supervised,yan2013learn,feng2015cascaded}, random ferns~\cite{cao2014face,ren2014face,zhu2016unconstrained} and even deep neural networks~\cite{zhu2016face,Trigeorgis_2016_CVPR,Zhang_2016_CVPR,Jourabloo_2016_CVPR}.
However, cascaded regression is usually based on a non-parametric Point Distribution Model (PDM) and merely recovers face shapes.
In this paper, we extend cascaded regression to our proposed UT-AAM fitting that recovers the shape and texture information jointly.
Moreover, we demonstrate the capacity of the proposed model to synthesise realistic 2D face instances for learning-based facial landmark detector training.

\section{Overview of tensor-based AAM}
\label{Sec_TAAM}
In this paper, scalars, vectors, matrices and higher-order tensors are 
denoted by lower-case letters ($a$, $b$, ...), bold lower-case letters ($\mathbf{a}$, $\mathbf{b}$, ...), bold upper-case letters ($\mathbf{A}$, $\mathbf{B}$, ...) and calligraphic upper-case letters ($\mathcal{A}$, $\mathcal{B}$, ...) respectively.

\subsection{Active Appearance Model (AAM)}
The classical AAM has two PCA-based parametric models, \ie shape and texture models.
For a 2D face image, the face shape $\mathbf{s} = [x_1, y_1, ..., x_{L}, y_{L}]^T$ is a vector formed by concatenating the 2D coordinates of $L$ pre-defined key points.
Given a set of annotated face images, a PCA-based shape model can be obtained:
\begin{equation}
 \label{shapemodel}
 \mathbf{s} = \bar{\mathbf{s}}  + \sum_{k=1}^{N_s} \alpha_k \mathbf{s}_k,
\end{equation}
where $\bar{\mathbf{s}} $ is the mean shape, $\mathbf{s}_k$ is the $k$th eigenvector obtained by applying PCA to training shapes aligned with Procrustes analysis, and $\alpha_k$ is the corresponding model parameter.

To obtain a texture model, pixels inside the face shape of a training image are first warped to a reference shape, \eg the mean shape, using a piece-wise affine transformation~\cite{Matthews2004}.
Then raster-scanning is applied to convert the 2D texture of a warped face to a texture vector $\mathbf{t} \in \mathbb{R}^{I_t}$.
Last, AAM applies PCA to construct a texture model:
\begin{equation}
 \label{appmodel}
\mathbf{t} = \bar{\mathbf{t}} +\sum_{k=1}^{N_t}\beta_k\mathbf{t}_k,
\end{equation}
where $\bar{\mathbf{t}} $ is the mean texture, $\mathbf{t}_k$ is the $k$th 
eigenvector obtained by PCA and $\beta_k$ is the corresponding model parameter.

The classical PCA-based AAM is capable of representing the majority of shape and texture variations observed in a training dataset.
These variations are parametrised by the coefficients of shape and texture models, \textit{i.e.} $\boldsymbol{\alpha}$ and $\boldsymbol{\beta}$. 
Given a new face image $\mathbf{I}$, AAM can reconstruct and model the shape and texture information of the face using a fitting algorithm.
The goal of AAM fitting is to adjust model parameters to minimise the pixel intensity difference between a generated face instance and an input image:
\begin{equation}
\label{FUN:AAM_Fitting_Goal}
\|
\bar{\mathbf{t}} 
+ 
\sum_{k=1}^{N_t}\beta_k \mathbf{t}_k
-
W(\mathbf{I}, \boldsymbol \alpha) 
\|_2^2,
\end{equation}
where $W(\mathbf{I}, \boldsymbol \alpha)$ is a function that warps the face texture inside the shape generated by the shape model with parameter $\boldsymbol{\alpha}$ to the reference shape.
This non-linear optimisation problem can be solved using either gradient-descent-based or learning-based approaches~\cite{Matthews2004}. 

\subsection{Tensor-based AAM}
\subsubsection{Higher-order Singular Value Decomposition}
Tensors are higher-order extensions of vectors and matrices.
An $N$th-order tensor $\mathcal{X} \in \mathbb{R}^{I_1 \times I_2 \times ... \times I_N}$ is an $N$-dimensional array with multiple indices.
Given a face dataset with $I_i$ identity, $I_p$ pose, $I_e$ expression and $I_l$ illumination variations, the shape or texture information of the dataset can naturally be expressed as a tensor.
In the shape tensor $\mathcal{S} \in \mathbb{R}^{I_i \times I_p \times I_l \times I_e \times I_s}$,
the element $s_{(i_i, i_p, i_l, i_e, i_s)}$ denotes the $i_s$th entry of the face shape vector for the $i_i$th identity with the $i_p$th pose, $i_l$th illumination and $i_e$th expression states, where $I_s = 2L$ is the dimensionality of a face shape vector.
Similarly, the texture tensor $\mathcal{T} \in \mathbb{R}^{I_i \times I_p \times I_l \times I_e \times I_t}$ reorganises the texture vectors of a training dataset in a tensor fashion.

In contrast to the PCA method used in AAM, T-AAM uses a multilinear subspace analysis to construct tensor-based shape and texture models.
To this end, tensor decomposition algorithms are used~\cite{tucker1966some,carroll1970analysis,harshman1970foundations,kolda2009tensor}. 
The two most important and popular tensor decomposition methods are CANDECOMP/PARAFAC (CP)~\cite{carroll1970analysis,harshman1970foundations} and Tucker~\cite{tucker1966some} tensors.
In one work, T-AAM uses Tucker tensor decomposition to obtain shape and texture models.

Given an $N$th-order tensor $\mathcal{X}$, Tucker tensor decomposition results in:
\begin{equation}
\label{FUN: Tucker_Decom}
\mathcal{X}=\mathcal{C}\times_1 \mathbf{U}_1\times_2 \mathbf{U}_2 ...\times_N \mathbf{U}_N,
\end{equation}
where $\mathcal{C}\in \mathbb{R}^{I_1 \times I_2 \times...\times I_N}$ is the core tensor
with the same dimensionality of the input tensor $\mathcal{X}$, which models the interaction between the orthonormal mode matrices $\mathbf{U}_n \in \mathbb{R}^{I_n \times I_n} (n = 1, ..., N)$.
The mode-n product `$\times_n$' between a tensor $\mathcal{X}$ and a matrix $\mathbf{Y} \in \mathbb{R}^{J \times I_n}$ results in a new tensor
$\mathcal{Z} \in \mathbb{R}^{I_1 \times ... \times J \times ... \times I_N}$, in which each element is calculated by:
\begin{equation}
\label{FUN:mode-n}
z_{(i_1, ..., i_{n-1},j,i_{n+1},...,i_N)} = \sum_{i_n = 1}^{I_n} x_{(i_1,...,i_N)}y_{(j,i_n)}.
\end{equation}

To perform Tucker tensor decomposition, Higher Order Singular Value Decomposition (HOSVD) is usually used, which is also known as the Tucker-1 tensor decomposition method~\cite{de2000multilinear,tucker1966some}.
HOSVD calculates the mode-n matrix $\mathbf{U}_n$ using the left singular matrix of SVD decomposition to the mode-n unfolded matrix $\mathbf{X}_{(n)} \in \mathbb{R}^{I_n \times I_1...I_{n-1}I_{n+1}...I_N}$ of $\mathcal{X}$.
To unfold a tensor along the $n$th mode, we reorder all the entries in the tensor by stacking the vectors along the $n$th mode as column vectors in a matrix. 
Last, the core tensor $\mathcal{C}$ is obtained by:
\begin{equation}
\label{FUN: coretensor}
\mathcal{C}=\mathcal{X}\times_1 \mathbf{U}_1^T\times_2 \mathbf{U}_2^T ...\times_N \mathbf{U}_N^T.
\end{equation}

\subsubsection{Constructing T-AAM}
Given a shape tensor $\mathcal{S}$, the use of HOSVD results in:
\begin{equation}
\label{Equ_shape_tensor_decomposition}
\mathcal{S} = \mathcal{C}_s \times_1\mathbf{S}_i\times_2\mathbf{S}_p \times_3
\mathbf{S}_l\times_4 \mathbf{S}_e\times_5 \mathbf{S}_s,
\end{equation}
where $\mathcal{C}_s \in \mathbb{R}^{I_i \times I_p \times I_l \times I_e \times I_s}$ is the shape core tensor,
$\mathbf{S}_i \in \mathbb{R}^{I_i \times I_i}, \mathbf{S}_p \in \mathbb{R}^{I_p \times I_p}, \mathbf{S}_l\in \mathbb{R}^{I_l \times I_l},$ $\mathbf{S}_e \in \mathbb{R}^{I_e \times I_e}$ and $\mathbf{S}_s\in \mathbb{R}^{I_s \times I_s}$ are mode matrices representing the decomposed identity, pose, illumination, expression and shape subspaces.
In the same manner, a texture tensor $\mathcal{T}$ is decomposed as:
\begin{equation}
\label{Equ_texture_tensor_decomposition}
\mathcal{T}  =  \mathcal{C}_t \times_1\mathbf{T}_{i}\times_2\mathbf{T}_{p}\times_3
\mathbf{T}_{l}\times_4 \mathbf{T}_{p}\times_5 \mathbf{T}_{t},
\end{equation}
in which the elements have similar meanings as those in the shape tensor decomposition.

As a 2D face shape is mainly influenced by pose and expression variations, T-AAM constructs a shape basis sub-tensor:
\begin{equation}
\label{Equ_Basis_Shape_Tensor}
\mathcal{B}_s  =  \mathcal{C}_s \times_2 \mathbf{c}_p^{T}\mathbf{S}_{p} \times_4 \mathbf{c}_e^{T}\mathbf{S}_{e} \times_5 \mathbf{S}_{s},
\end{equation}
where $\mathbf{c}_p \in \mathbb{R}^{I_p}$ and $\mathbf{c}_e \in \mathbb{R}^{I_e}$ are pose and expression mixture coefficient vectors that indicate a linear combination of different pose or expression variation states, satisfying $\sum_{k=1}^{I_p}{c_p(k)} = 1$, $\sum_{k=1}^{I_e}{c_e(k)} = 1$, $ 0 \leqq c_p(k) \leqq 1$ and $ 0 \leqq c_e(k) \leqq 1$.

Because the pose and expression variations have already been considered in the shape basis sub-tensor, a texture basis sub-tensor is obtained by:
\begin{equation}
\label{Equ_Basis_Texture_Tensor}
\mathcal{B}_t 
=
 \mathcal{C}_t \times_3 \mathbf{c}_l^{T}\mathbf{T}_{l} \times_5 \mathbf{T}_{t},
\end{equation}
where $\mathbf{c}_l \in \mathbb{R}^{I_l}$ is an illumination mixture coefficient vector that indicates a linear combination of different illumination variation states, satisfying $\sum_{k=1}^{I_l}{c_l(k)} = 1$ and $ 0 \leqq c_l(k) \leqq 1$.

Last, T-AAM constructs a variation-specific shape model:
\begin{equation}
\label{Equ_tensor_shape_model}
\mathbf{s} = \bar{\mathbf{s}} (\mathbf{c}_{p},\mathbf{c}_{e}) +\sum_{k=1}^{N_s}{\alpha_k \mathbf{s}_k(\mathbf{c}_{p},\mathbf{c}_{e})},
\end{equation}
where $\bar{\mathbf{s}} (\mathbf{c}_{p},\mathbf{c}_{e})$ is the weighted mean shape computed over all training shapes using the pose and expression mixture coefficient vectors,
$\mathbf{s}_k(\mathbf{c}_{p},\mathbf{c}_{e})$ is the $k$th column vector of the unfolded matrix $\mathbf{B}_s \in \mathbb{R}^{I_{s} \times I_{i} I_{l}}$ of the shape basis tensor along the $5$th mode and $\alpha_k$ is the corresponding model parameter.
Similarly, a variation-specific texture model is constructed by T-AAM:
\begin{equation}
\label{Equ_tensor_texture_model}
\mathbf{t} = \bar{\mathbf{t}}(\mathbf{c}_l) +\sum_{k=1}^{N_t}{\beta_k \mathbf{t}_k(\mathbf{c}_l)},
\end{equation}
where $\bar{\mathbf{t}} (\mathbf{c}_l)$ is the weighted mean texture computed over all training texture vectors using the illumination mixture coefficient vector,
$\mathbf{t}_k(\mathbf{c}_l)$ is the $k$th column vector of the unfolded matrix $\mathbf{B}_t \in \mathbb{R}^{I_{t} \times I_iI_pI_e}$ of the texture basis tensor along the $5$th mode and $\beta_k$ is the corresponding model parameter.

\subsubsection{T-AAM fitting}
Given a new image, T-AAM first estimates the states of pose, expression and illumination of the face in the image.
T-AAM applies face and eye detection algorithms to perform rigid face alignment.
Then a tensor-based prediction method is used to estimate the mixture coefficient vectors $\mathbf{c}$ for pose, expression and illumination.
According to the constraint used for the value of each element in a mixture coefficient vector, T-AAM can be divided into discrete or continuous T-AAM.
The assumption of discrete T-AAM is that the state of a variation mode of a new face belongs to one of the states of the corresponding variation mode in the training dataset.
Hence the value of each element in a mixture coefficient vector can only be either 0 or 1, \ie $c \in \{0,1\}$.
In fact, the state of a specific variation mode of a face could be a linear combination of many discrete variations states in the training dataset.
In such a case, continuous T-AAM sets the value of each element in a mixture coefficient vector as a continuous variable, \ie $c \in [0,1]$.

For discrete T-AAM, we can pre-compute a set of variation-specific models offline and select the best matched one during online fitting.
In contrast, continuous T-AAM generates variation-specific shape and texture models online during model fitting.
Both of them rely on the estimation step in the fitting phase, \ie obtaining the mixture coefficient vector of each variation mode by estimating the states of pose, expression and illumination modes. 
Given the estimated mixture coefficient vectors, the corresponding variation-specific shape and texture models are selected to fit the input image using a gradient-descent-based optimisation approach~\cite{Lee2009}.

Compared with AAM, the multilinear subspace analysis used in T-AAM decouples shape and texture information into different variation-related subspaces, and provides a structured and compact representation of the shape and texture information of a dataset.
For model fitting, T-AAM creates variation-specific models by estimating the variation states of a new face.
This benefits the gradient-descent-based fitting algorithm in two ways.
On one hand, the use of variation-specific models initialises model parameters closer to the global minima.
On the other hand, by fixing variation types, the search space shrinks to a smaller subspace that is easier to handle.
In fact, T-AAM generates a variation-specific model using only the shape and texture eigenvectors related to a specific variation type,
hence it fits a new face exhibiting the same variation states much more readily.

\section{The proposed UT-AAM framework}
\label{Sec_UTAAM}
Although the tensor algebra has the capacity to support multiple-factor data analysis, and T-AAM has been reported to be a powerful approach for dealing with appearance variations in face analysis, the use of T-AAM is not without difficulties.
The major issue is that the classical T-AAM is a collection of many variation-specific models.
For T-AAM fitting, the estimation of variation states of a new face is in practice hard and restricted by the accuracy of face and eye detectors.
In addition, T-AAM cannot deal with the problems of self-occlusion and missing training samples.
To address these issues, we propose a unified framework for tensor-based AAM.
To achieve unification, the proposed UT-AAM framework introduces four techniques.
First, UT-AAM creates a unified tensor model across different variation modes.
Second, to deal with the problem of self-occlusion of large-scale pose variations, a uniform representation strategy is advocated.
Third, the proposed UT-AAM method can be created from an incomplete training dataset with missing training samples by introducing tensor completion approaches.
Last, a new cascaded-regression-based model fitting algorithm is presented, which does not require estimating the variation states for fitting a new face.

\subsection{Unified shape and texture models}
Unlike the variation specific shape and texture models used in the classical T-AAM method, UT-AAM constructs a single tensor-based shape or texture model.
Given a set of annotated face images, the corresponding shape tensor $\mathcal{S} \in \mathbb{R}^{I_{i} \times I_{p} \times I_{l} \times I_{e} \times I_{s}}$ and texture tensor $\mathcal{T} \in \mathbb{R}^{I_{i} \times I_{p} \times I_{l} \times I_{e} \times I_{t}}$, we apply HOSVD to the shape and texture tensors for Tucker tensor decomposition.
The resulting unified tensor-based shape model is given as:
\begin{equation}
\mathbf{s} 
=
\bar{\mathbf{s}} 
+
\mathcal{C}_s 
\times_1
\mathbf{a}_i^{T}\mathbf{S}_{i}
\times_2
\mathbf{a}_p^{T}\mathbf{S}_{p}
\times_3
\mathbf{a}_l^{T}\mathbf{S}_{l}
\times_4 
\mathbf{a}_e^{T}\mathbf{S}_{e}
\times_5
\mathbf{S}_{s}.
\end{equation}
Because the shape of a subject is independent of illumination variations, we can compress the shape model as:
\begin{equation}
\mathbf{s} 
=
\bar{\mathbf{s}} 
+
\tilde{\mathcal{C}}_s 
\times_1
\mathbf{a}_i^{T}\mathbf{S}_{i}
\times_2
\mathbf{a}_p^{T}\mathbf{S}_{p}
\times_4 
\mathbf{a}_e^{T}\mathbf{S}_{e},
\end{equation}
where $
\tilde{\mathcal{C}}_s 
= 
\mathcal{C}_s
\times_3
\mathbf{S}_{l}
\times_5
\mathbf{S}_{s}$.
In addition, we have to apply a global affine transform $G(\mathbf{s}, \mathbf{p}_g)$ to the shape,
where $\mathbf{p}_g = [s, \theta, t_x, t_y]^T$ is the global affine transform parameter controlling scale, rotation and translation.
In this unified tensor-based shape model, a new shape can be represented by a long parameter vector $\mathbf{p} = [\mathbf{p}_g^T, \mathbf{a}_i^T, \mathbf{a}_p^T, \mathbf{a}_e^T ]^T$.

Similarly, we can obtain a unified texture model:
\begin{equation}
\mathbf{t} 
=
\bar{\mathbf{t}} 
+
\tilde{\mathcal{C}}_t 
\times_1
\mathbf{b}_i^{T}\mathbf{T}_{i}
\times_2
\mathbf{b}_p^{T}\mathbf{T}_{p}
\times_3
\mathbf{b}_l^{T}\mathbf{T}_{l}
\times_4 
\mathbf{b}_e^{T}\mathbf{T}_{e},
\end{equation}
where $
\tilde{\mathcal{C}}_t
= 
\mathcal{C}_t
\times_5 
\mathbf{T}_{t}$.
The texture of a face can be expressed by the texture model parameter vector $\mathbf{q} = [\mathbf{b}_i^T, \mathbf{b}_p^T, \mathbf{b}_l^T, \mathbf{b}_e^T ]^T$.
Last, a new face instance can be represented by a unified parameter vector concatenating the shape and texture model parameters $[\mathbf{p}^T, \mathbf{q}^T]^T$.
For a new face image, the goal of UT-AAM fitting is to find the model parameter vector best representing the input face.

\subsection{Uniform landmarking strategy for self-occluded faces}
\label{ulandmark}
As discussed at the beginning of this paper, another practical issue for building a unified AAM is the problem of self-occlusion posed by large-scale head rotations.
Extreme pose variation of a face often results in some facial parts being invisible. 
However, classical T-AAM does not consider the problem of self-occlusion for extreme pose variations.
This is a crucial issue for constructing unified shape and texture models,
as T-AAM would require consistent representation for a face shape or texture across large-scale pose variations, \textit{i.e.} having the same number of face key points and the same dimensionality of extracted texture vectors.
\begin{figure}[!t]
\centering
 \includegraphics[trim = 0mm 40mm 0mm 0mm, clip, width=.8\linewidth]{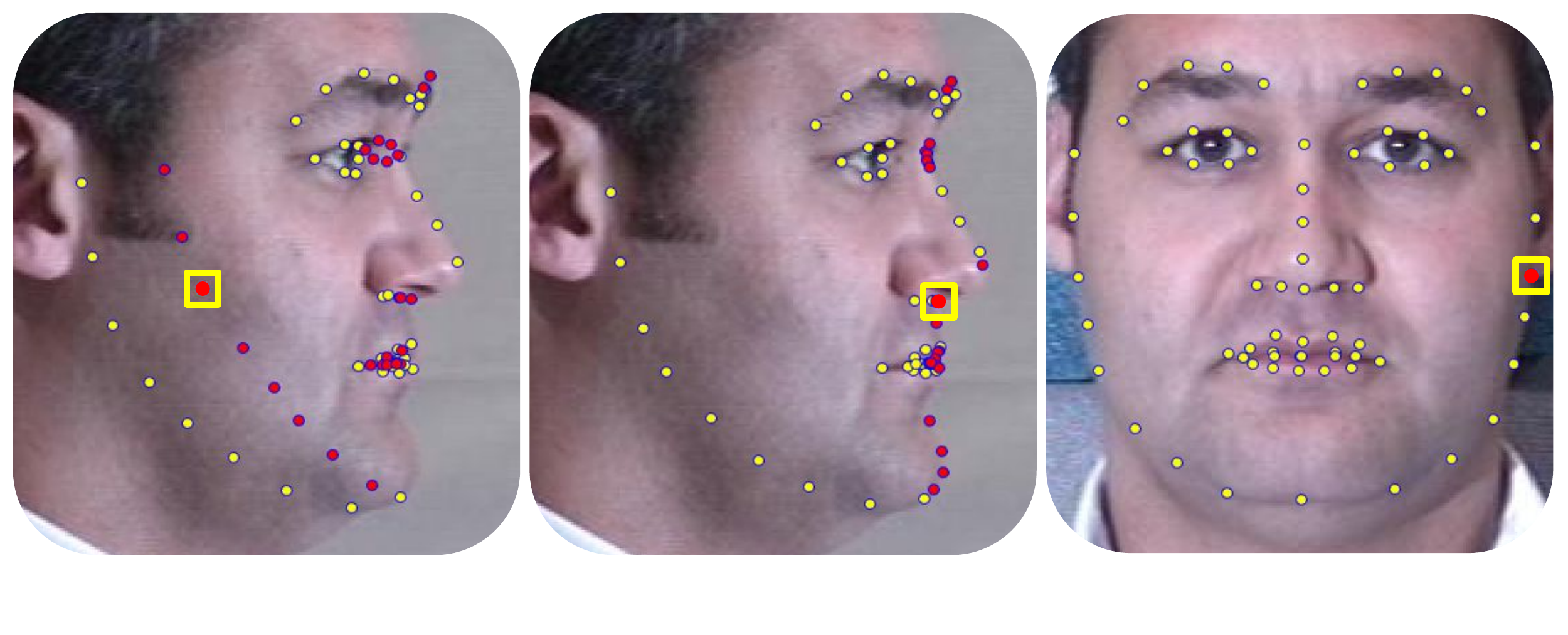}
\caption{Different definitions for self-occluded facial landmarks. \textit{Left}: using 3D face models; \textit{Middle}: the proposed strategy; \textit{Right}: the corresponding frontal face.}
 \label{fig_2}
\end{figure}

To mitigate this problem, two different strategies have been developed in previous work.
The first one is to use view-specific models, in which only visible facial parts are considered.
For example, view-based AAM builds 5 different AAMs for the viewpoints at $-90^\circ$, $-45^\circ$, $0^\circ$, $45^\circ$ and $90^\circ$, using different definitions of face key points~\cite{Cootes2000}.
However, this strategy is not feasible for our UT-AAM because it requires consistent shape/texture vectors across poses.
Another solution is to use 3D face models.
Feng \etal used the ground-truth of self-occluded face key points by projecting 3D face vertices to 2D images~\cite{feng2015cascaded}.
The same strategy has also been used in 3D-assisted 2D facial landmark detection~\cite{zhu2016face}.
However, it is very hard to manually annotate/guess such kind of face key points, as the red points shown in the left column of Fig.~\ref{fig_2}.
We have to build a 3D face model and fit it to 2D face images to obtain self-occluded face key points.
Both the construction and fitting procedures of a 3D face model are hard tasks to implement.

In contrast to these two approaches, we propose a uniform landmarking strategy that is simple to implement and easy for a human to annotate self-occluded facial landmarks.
In our UT-AAM, a key point of invisible facial parts is re-defined as the cross point of the face outline and the horizontal line passing the occluded key point, as illustrated at the middle column of Fig.~\ref{fig_2}.
This simple landmarking strategy guarantees the same dimensionality of face shapes for all pose variations. 
For face texture, the reference shape is the average value over all frontal faces and the Delaunay triangulation is used to build the corresponding face mesh.
Then all shapes with different viewpoints share the same triangulation results derived from the reference shape.
Last, shape-free face patches are obtained by piece-wise affine warp that maps the global texture inside an original face shape to the reference shape~\cite{Matthews2004}. 
The proposed universal landmarking strategy ensures the same representation of face shapes regardless of pose angles, so that we can obtain a single reference shape and a consistent reference mesh for extracting uniform face texture vectors.

\subsection{Coping with missing training samples}
In practice, one obstacle impeding the use of a traditional tensor-based model is the problem of missing training samples.
Sometimes it is not feasible to collect a complete dataset with all possible combinations of variations, which results in an incomplete training set.
For example, some subjects may fail to participate in a particular data collection session.
Thus it is clearly important to be able to build a tensor-based model from an incomplete training dataset and to investigate how this model performs in practice.

As a tensor can be unfolded to a matrix, any matrix completion method dealing with missing values can be used for tensor completion~\cite{vlasic2005face}, in which missing entries are usually randomly distributed in a matrix.
However, a missing training sample in our case leads to the whole column vectors of the shape and texture matrices being missing.
In addition, matrix completion methods do not consider the interaction and structure of the signals in a dataset.
In this section, to address the problem of missing training samples, we use two tensor completion methods , \ie the CP-based CP-WOPT~\cite{acar2011scalable} and Tucker-based M$^2$SA algorithms~\cite{Geng2011}.

Given an incomplete tensor $\mathcal{X} \in \mathbb{R}^{I_1 \times \cdots \times I_N}$ with missing entries,
the goal of a tensor completion algorithm is to find a complete tensor $\mathcal{X}'$ to minimise
$\lVert \mathcal{X}' - \mathcal{X}^*\rVert$,
where $\mathcal{X}^*$ is the ground truth tensor without missing entries.
`$\lVert \rVert$' is the norm of a tensor, which is defined as the square root of the sum of the squares of all the elements in the tensor.
However, this complete ground truth tensor is unknown in practice;
hence M$^2$SA modifies the cost to find the best low-rank approximation $\mathcal{X}'$ for available entries by minimising
\begin{equation}
\label{Equ_tensor_completion_objective}
\lVert
\mathcal{O} \ast (\mathcal{X}-\mathcal{X}')
\rVert,
\end{equation}
where `$\ast$' is the entry-wise product of two tensors.
$\mathcal{O} \in \mathbb{R}^{I_1 \times \cdots \times I_{N}}$
is an index tensor with the same size as $\mathcal{X}$, in which the value of an entry is set to 1 (or 0) when it is available (or missing).
Given an initial estimate of $\mathcal{X}'$, M$^2$SA applies a power method that iteratively updates $\mathcal{X}'$ to solve the above optimisation problem.
In contrast, CP-WOPT treats the task as a weighted least squares problem and uses a first-order optimisation approach to solve it.
For more details of these two tensor completion algorithms, the reader is referred to~\cite{Geng2011} and~\cite{acar2011scalable}. 

To use the M$^2$SA and CP-WOPT methods, we have to first initialise the missing values for an incomplete tensor.
A straightforward initialisation method is to use random values or the average value of some other available entries.
However, these approaches do not consider the variations of a specific missing entry.
A better way is to predict the missing entry using only the available entries that share the same variability types.
To simplify the discussion, we take the shape tensor as an example to introduce the proposed initialisation algorithm.
Note that the algorithm discussed below is also applicable to texture tensors.

Given an incomplete shape tensor $\mathcal{S} \in \mathbb{R}^{I_{i} \times I_{p} \times I_{l} \times I_{e} \times I_{s}}$ with missing entries, we first initialise a missing entry using the average value of all available entries with the same pose, illumination and expression variations:
\begin{equation}
\label{equ_and}
\mathbf{s}^m_{i_{i}, i_{p}, i_{l}, i_{e},:}
\leftarrow
\frac {\sum_{(i'_{p}=i_{p}) \wedge (i'_{l} = i'_{l}) \wedge (i'_{e}=i_{e}) } {\mathbf{s}^a_{i'_{i},i'_{p}, i'_{l}, i'_{e}, :}}}{N^a},
\end{equation}
where: the superscripts `$m$' and `$a$' stand 
for missing entry and available entry respectively; 
the subscripts stand for the position of the corresponding entry in $\mathcal{S}$;
and $N^a$ is the number of all the available entries with the same variations as the missing one.

It is obvious that the constraint in equation~(\ref{equ_and}) is very strong.
The `AND' operator used in the initialisation method has only one free factor `identity', whereas all the other factors are fixed. 
Thus the initialised missing entries do not contain other types of variations.
However, we may not have enough available entries with the same variation modes as the missing one, especially when the proportion of missing entries is very high.
This problem can be solved by replacing the `AND' operator with the `OR' operator:
\begin{equation}
\label{equ_or}
\mathbf{s}^m_{i_{i}, i_{p}, i_{l}, i_{e}, :}
\leftarrow
\frac {\sum_{(i'_{p}=i_{p}) \vee (i'_{l} = i'_{l}) \vee (i'_{e}=i_{e})} {\mathbf{s}^a_{i'_{i},i'_{p}, i'_{l}, i'_{e}, :}}}{N^a},
\end{equation}
in which the `OR' operator makes it easier to find enough available entries sharing the variations of a missing entry.

In summary, for a missing entry, we first use the `AND' operator for initialisation.
If no available entries exists, we switch to using the `OR' operator.
However, in some extreme cases, even the use of `OR' operator may not find an available entry to initialise the missing one.
In such a case, we initialise the missing entry with a random value in $[0,1]$.

\subsection{Cascaded regression for UT-AAM fitting}
Given a new face image, the classical T-AAM first applies a tensor-based estimation algorithm to predict the states of pose, expression and illumination variations of the face~\cite{Lee2009}.
Then a gradient-descent-based model fitting algorithm is used to fit the generated variation-specific models to the image.
This pipeline has two main drawbacks.
The first and most important one is that the estimation step highly relies on the accuracy of a face or eye detector, which cannot be guaranteed, especially for faces exhibiting extreme appearance variations.
With an inaccurate estimation result, the generated variation-specific shape and texture models cannot be well fitted to the image.
The second drawback is that the fitting algorithm is gradient-descent-based, hence can easily be trapped in local minima.

To deal with the above issues, we propose a new fitting algorithm for our UT-AAM, using a learning-based, coarse-to-fine framework, \ie cascaded regression~\cite{dollar2010cascaded,xiong2013supervised}.
Given a face image $\mathbf{I}$ and the initial model parameter vector $\mathbf{p}$, the aim of a regression method is to construct a mapping function:
\begin{align}
&\phi:  \hspace{0.1cm} f(\mathbf{I},\mathbf{p})  \mapsto \delta \mathbf{p},\\
&s.t. \hspace{0.1cm} \| \mathbf{p} + \delta \mathbf{p}  -  \mathbf{p}^{\ast} \|^2_2 = 0, \nonumber
\end{align}
where $f(\mathbf{I},\mathbf{p}) \in \mathbb{R}^{N_f}$ is a feature extraction function that is related to the current model parameter vector, $N_f$ is the dimensionality of an extracted feature vector, $\delta\mathbf{p}$
is the update to the current model parameter vector and $\mathbf{p}^*$ is the ground truth parameter vector of the face.
Given a set of training examples, we can learn this mapping function by any regression method, \eg linear regression, random ferns or even deep neural networks.
However, a single regressor will not handle the task very well.
To address this issue, cascaded regression constructs a strong regressor by cascading $M$ weak regressors, $\Phi = \{\phi_1, ..., \phi_M\}$.

To construct these cascaded weak regressors, we first train the first weak regressor using the original training samples.
Then we apply the first trained weak regressor to update all the initial model parameters, $\mathbf{p} \leftarrow \mathbf{p}+\delta\mathbf{p}$, for the second weak regressor training.
The required number of weak regressors can be trained by iteratively repeating this procedure.
In this paper, each weak regressor is a linear regressor, \ie $\phi_m: \delta \mathbf{p} = \mathbf{A}_m f(\mathbf{I},\mathbf{p}) + \mathbf{b}_m$, where $\mathbf{A}_m \in \mathbb{R}^{N_p \times N_f}$ is the projection matrix, $\mathbf{b}_m \in \mathbb{R}^{N_p}$ is the offset and $N_p$ is the dimensionality of a parameter vector $\mathbf{p}$. 
For the training of the $m$th weak regressor, the cost function is:
\begin{equation}
\underset{\mathbf{A}_m,\mathbf{b}_m}{\mathrm{argmin}}
\sum_{n=1}^{N}
{
\lVert 
\mathbf{A}_m f(\mathbf{I}_n,\mathbf{p}_n)
+
\mathbf{b}_m
-
\delta\mathbf{p}_n
\rVert^2_2
}
+
\lambda
\lVert
\mathbf{A}_m
\rVert^2_F
,
\end{equation}
where $f(\mathbf{I}_n,\mathbf{p}_n)$ is the extracted feature vector of the $n$th training example, $\delta\mathbf{p}_n = \mathbf{p}_n^* - \mathbf{p}_n$ is the difference between the current model parameter and the ground truth model parameter, $\lambda$ is the weight of the regularisation term and $\lVert * \rVert_F$ is the Frobenius norm for a matrix.
It should be noted that $\mathbf{p}_n$ and $\delta\mathbf{p}_n$ are updated after each weak regressor training.
\begin{algorithm}[t]
\begin{algorithmic}[1]
\vspace{0.03in}
\STATE
\textbf{input}
An image $\mathbf{I}$, the trained cascaded regressors $\Phi = \{\phi_1, ..., \phi_M\}$, initialised model parameters $\mathbf{p}$ and $\mathbf{q}$;

\FOR{$m = 1$ to $M$}
\STATE Extract local features $f(\mathbf{I}, \mathbf{p})$;

\STATE Apply the $m$th weak regressor $\phi_m$ to obtain $\delta\mathbf{p}$;

\STATE Update the current model parameter $\mathbf{p} \leftarrow \mathbf{p} + \delta\mathbf{p}$;
\ENDFOR

\STATE Warp the texture in the current shape estimate to the reference shape and estimate $\mathbf{q}$;

\RETURN Predicted model parameters $\mathbf{p}$ and $\mathbf{q}$.
\vspace{0.03in}
\end{algorithmic}
\caption{The proposed T-AAM fitting algorithm}
\label{Algorithm_T-AAM_Fitting}
\end{algorithm}

Given a new face image, the initial parameter estimation $\mathbf{p}$ and a trained cascaded regressor $\Phi$,
the model parameter is iteratively updated using the weak regressors in $\Phi$.
In this paper, we extract HOG features around each face key point and concatenate them to a long vector as our extracted features, \ie $f(\mathbf{I}, \mathbf{p})$. 
Once we obtain the final estimate of the shape model parameter $\mathbf{p}$, the shape of a face can be calculated using our tensor-based shape model.
Then the global texture of the face image is obtained by wrapping the pixels in the estimated face shape to the reference shape,
and the texture model parameter $\mathbf{q}$ can be estimated as introduced in~\cite{Lee2009,lin2005tensor}.
The proposed fitting algorithm is summarised in Algorithm~\ref{Algorithm_T-AAM_Fitting}.

\section{Experimental results}
\label{Sec_Evaluation}
In this section, we first compare the proposed UT-AAM with the classical T-AAM on the Multi-PIE face dataset~\cite{gross2010multi}.
Then we demonstrate the capacity of the proposed UT-AAM to synthesise a large number of virtual faces and examine how these synthesised faces can improve the training of a facial landmark detector, using the 300-W face dataset~\cite{sagonas2016300}.

\subsection{Datasets and experimental settings}
\begin{figure}
\centering
\subfloat[Illumination]{
 \includegraphics[width=0.95\linewidth]{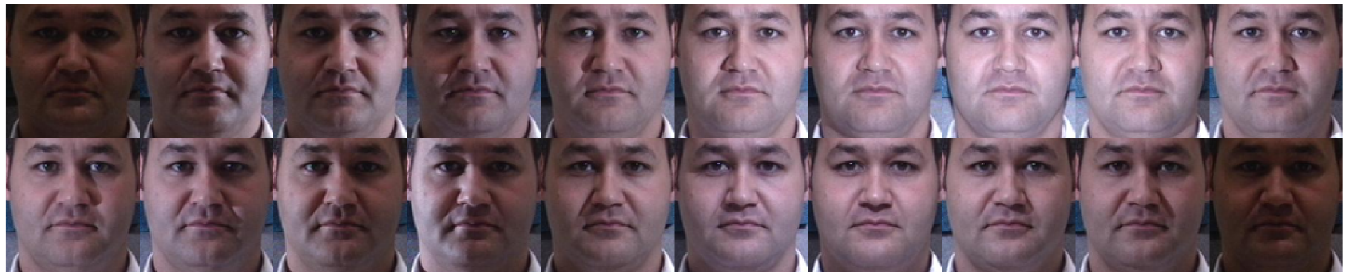}
}
\\
\subfloat[Pose]{
 \includegraphics[width=0.67\linewidth]{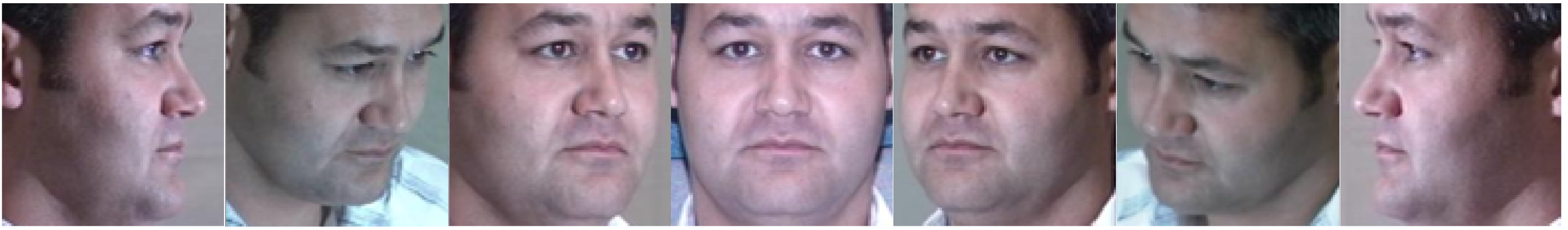}
}
\subfloat[Expression]{
 \includegraphics[width=0.29\linewidth]{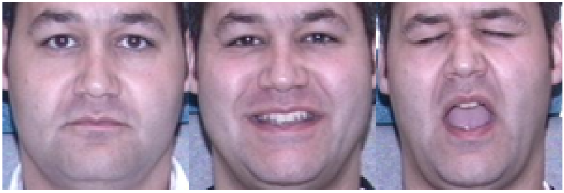}
}
\caption[Variations of the selected subset from Multi-PIE]{Variations of the selected subset from Multi-PIE.}
\label{fig_3}
\end{figure}

The Multi-PIE face dataset has more than 750000 images of 377 subjects, captured from 4 different sessions over the span of five months.
The images of a subject in Multi-PIE were captured across 15 poses, 20 lighting conditions and a range of expression
variations.
In our experiments, each face in Multi-PIE was manually annotated using 68 facial key points for model training and providing ground truth.
The locations of these 68 landmarks were defined as same as the 300-W face dataset.
For a self-occluded point, it was annotated using the landmarking strategy in Section~\ref{ulandmark}.
However, it is laborious to manually annotate all the Multi-PIE images; 
hence we only annotated a subset containing 60 subjects with 7 poses ($12\_0$, $08\_1$, $13\_0$, $05\_1$, $04\_1$, $19\_1$ and $01\_0$), 3 expressions (neutral from session-1, smile from session-3 and scream from session-4) and all the 20 illumination variations.
In total, 25200 images were manually annotated.
Fig.~\ref{fig_3} shows the variations of the subset.

The 300-W dataset has been widely used for benchmarking a facial landmark detection algorithm~\cite{sagonas2016300}.
For testing, 300-W contains 600 images (300 indoor and 300 outdoor), in which each face image has 68 landmarks.
For training, the dataset provides 68 landmarks for the XM2VTS, FRGC, AFW, HELEN and LFPW datasets.

The accuracy of different algorithms on Multi-PIE was measured in terms of the widely used point-to-point (pt-pt) error, \ie the average across all the landmarks of the Euclidean distance between the ground truth and fitted face shapes.
The Multi-PIE dataset was captured under a controlled scenario and the imaging parameters were fixed, hence the resolution and scale of all the images are unified.
However, for the 300-W dataset, the images were downloaded from the Internet so there is not consistency among face scales.
To address this issue, the pt-pt error normalised by inter-ocular distance of the face is used to measure the accuracy of a facial landmark detector on 300-W.


\subsection{UT-AAM versus T-AAM}
This part first compares our UT-AAM with the classical T-AAM, as well as the classical AAM, the view-based AAM (V-AAM) and the Fast Simultaneous Inverse Compositional (Fast-SIC) algorithm~\cite{tzimiropoulos2013optimization}.
Then we investigate the performance of our UT-AAM trained from an incomplete dataset with missing training samples.

\subsubsection{Model fitting}
In this experiment, we randomly selected 30 subjects with 12600 ($30 \times 20 \times3 \times7$) images from our annotated subset of Multi-PIE as the training set and the remaining 30 subjects with 12600 images were used as the test set.
We repeated this procedure 10 times and used the average pt-pt error to assess accuracy.
To initialise V-AAM and T-AAM, we assumed that the states of pose, expression and illumination of a test image were already known.
A benefit of our UT-AAM in practical applications is that we do not have to estimate the state of each variation type for a test face image.
For AAM, V-AAM and T-AAM, the gradient-descent-based Gauss-Newton fitting algorithm was used.
Fast-SIC is an advanced AAM fitting algorithm developed for unconstrained scenarios, which is based on the inverse compositional algorithm~\cite{tzimiropoulos2013optimization}.
For the proposed UT-AAM fitting algorithm, five linear regressors were cascaded.
\begin{table}[!t]
\renewcommand{\arraystretch}{1.2}
\centering
\caption{A comparison of different algorithms on the Multi-PIE face dataset in terms of fitting error and speed}
\label{table_1}
\begin{tabular}{lr@{$\pm$}lr@{$\pm$}lc}
\hline
Algorithm  & \multicolumn{2}{l}{Initial Error (pixel)} & \multicolumn{2}{l}{Fitting Error (pixel)}   & Speed (fps)\\    
\hline                                      
AAM  & 10.4143 & 0.1399  & 10.5387 & 0.2084    &      3\\
V-AAM & 5.0248 & 0.0657  & 3.1367 & 0.0745    &     3\\
T-AAM & 4.6182 & 0.1728  &  3.3438 & 0.1378    &      3\\
Fast-SIC  & 10.4143 & 0.1399  & 6.7840 & 0.2027    &   2  \\
UT-AAM & 9.4655 & 0.0374 &  \textbf{2.6964} & \textbf{0.0816} & \textbf{13}\\
\hline
\end{tabular}
\end{table}

The initial and fitting errors of different algorithms are shown in Table~\ref{table_1}.
It should be noted that both V-AAM and T-AAM require knowledge of the states of the pose, expression and illumination variation modes of a test image.
Then the corresponding variation-specific mean face is used to initialise the model fitting process.
This is the main reason why the initial errors of V-AAM and T-AAM are much lower than those of the classical AAM, Fast-SIC and our UT-AAM.
In contrast, UT-AAM does not require any prior information for model fitting.
In addition, gradient-descent-based fitting algorithms can be trapped by local minima when the initialisation is very far away from the global optimum, so that the fitting errors of AAM and Fast-SIC are very high.
In contrast, both V-AAM and T-AAM obtain much lower fitting errors even using the Gauss-Newton solver.
The success of V-AAM and T-AAM has two main origins.
The first is the use of variation-specific shape and texture models, which can be fitted to a new face image with the same variation states better than a generic model.
The second one, which may be more important, owes to the variation state estimation step that provides better initialisation for model fitting.
Note, the proposed UT-AAM with our learning-based model fitting algorithm beats all the other algorithms, including the state-of-the-art Fast-SIC, in terms of accuracy without requiring the estimation of the variation state of each test image.
Last, our UT-AAM fitting is much faster than the other algorithms with the speed of 13 fps (frames per second).
The speed was obtained on a Intel Xeon E5-2643 v3 CPU.

\begin{figure}
\centering
\includegraphics[trim = 6mm 25mm 4mm 15mm, clip, width=.85\linewidth]{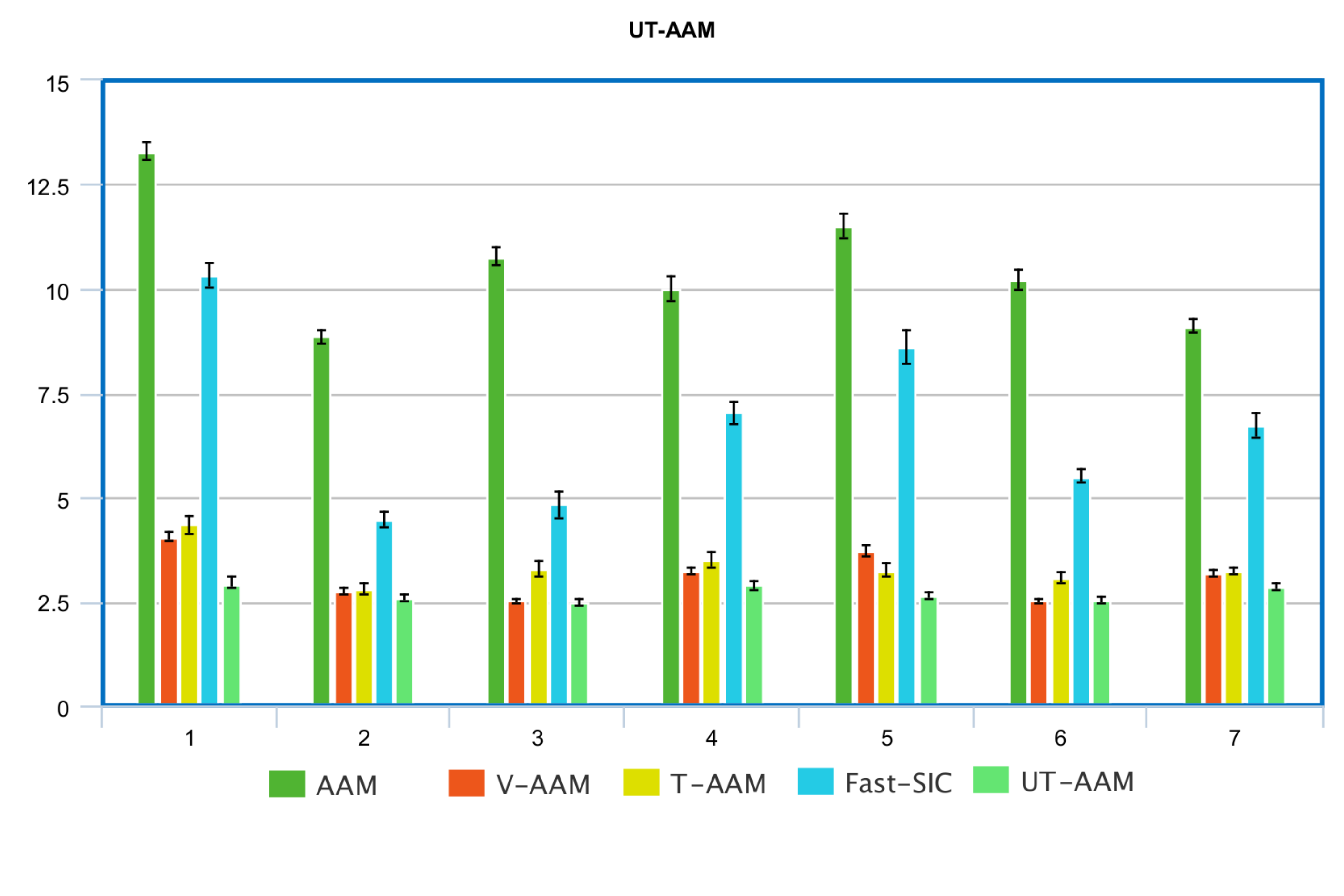}
\includegraphics[trim = -50mm 0mm 0mm 0mm, clip, width=0.8\linewidth]{./Img/fig3_2.png}
\caption{A comparison of different algorithms on the Multi-PIE face dataset parametrised by pose variations.}
\label{fig_4}
\end{figure}

To further investigate the performance of different algorithms, we present their fitting errors parametrised by pose variations in Fig.~\ref{fig_4}.
As can be seen in the figure, the classical AAM and Fast-SIC are incapable of fitting a new face with combined appearance variations in pose, expression and illumination, especially for faces with extreme pose variations (up to $\pm 90 ^\circ$ in yaw).
For V-AAM and T-AAM, as conjectured above, the use of variation-specific models provides a good mechanism for dealing with appearance variations.
However, the main drawback of both methods is their need for a separate step to estimate the state of variations for model initialisation, which is both time-consuming and in practice difficult for faces with a wide range of appearance variations.
In contrast, the proposed UT-AAM does not require the variation estimation step and outperforms all the other approaches.

\subsubsection{Dealing with missing training samples}
In this section, we evaluate the robustness of the proposed UT-AAM to missing training samples.
To this end, we first test the reconstruction accuracy of two tensor completion algorithms, \ie M$^2$SA and CP-WOPT, using both the random initialisation method and the proposed initialisation method.
Second, we evaluate the accuracy of the proposed UT-AAM in terms of model fitting error.
\begin{figure}
\centering
\subfloat[Shape]{
\label{fig_7_1}
 \includegraphics[trim = 40mm 93mm 44.5mm 95mm, clip, width=.7\linewidth]{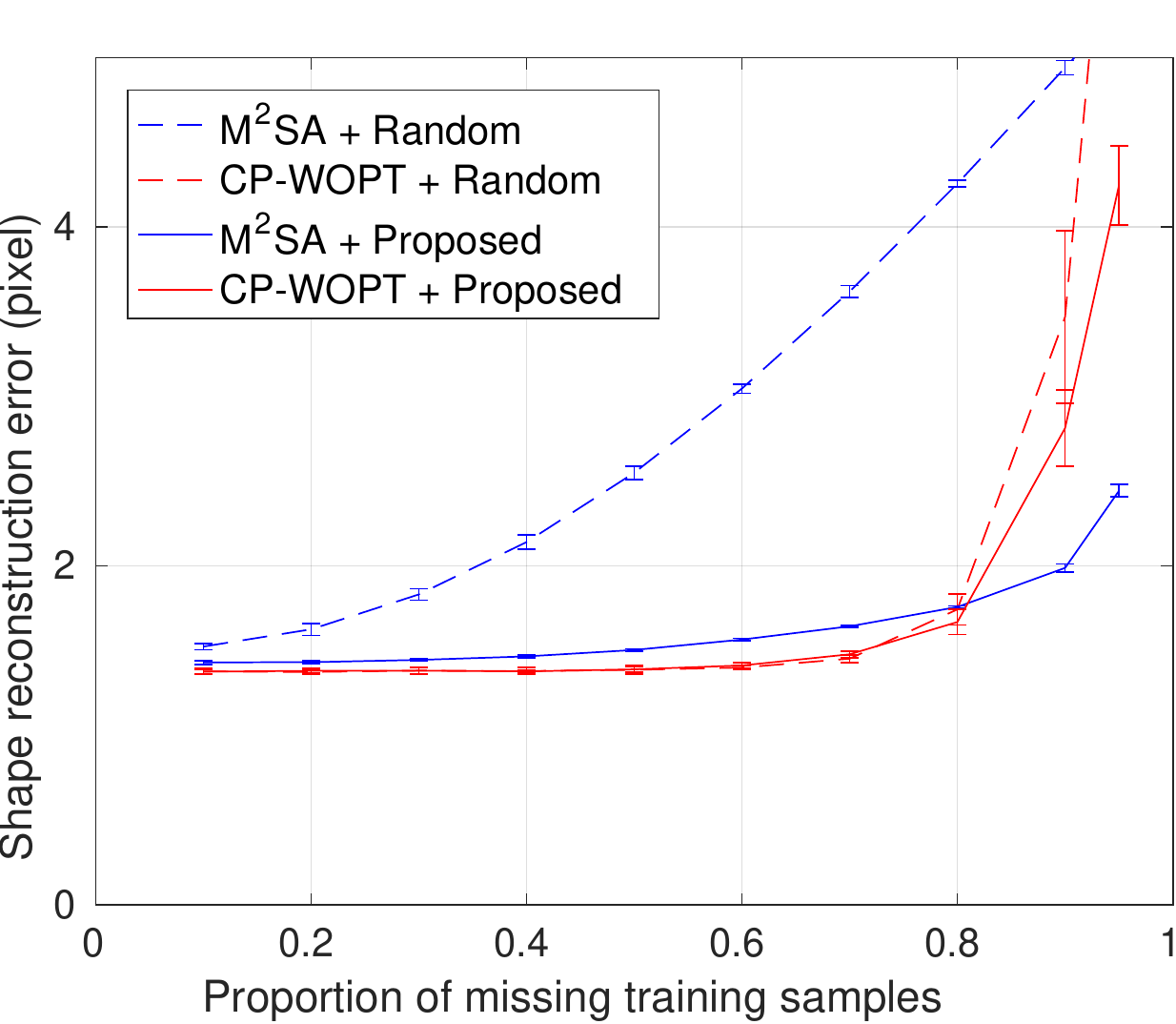}
}
\\
\subfloat[Texture]{
\label{fig_7_2}
 \includegraphics[trim = 37mm 93mm 44.5mm 95mm, clip, width=.7\linewidth]{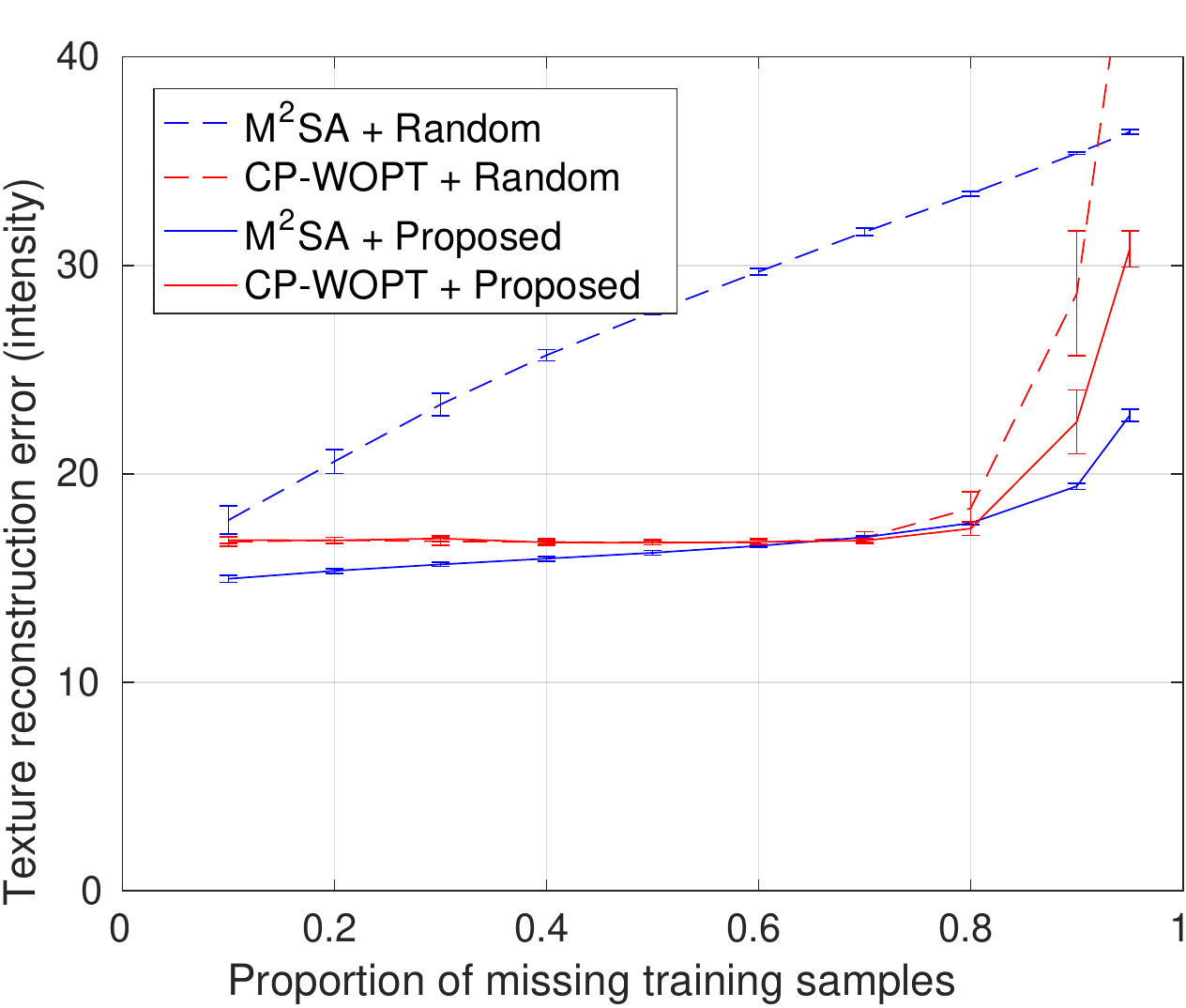}
}
\caption{A comparison of the accuracy of M$^2$SA and CP-WOPT tensor completion algorithms, parametrised by different proportions of missing training samples: (a) the shape reconstruction RMS errors; (b) the texture reconstruction RMS errors. Both the proposed and random initialisation methods are used.}
\label{fig_7}
\end{figure}

To evaluate the capacity of the M$^2$SA and CP-WOPT algorithms to reconstruct missing training samples of an incomplete tensor, we used the following root mean square (RMS) error as our performance criterion.
The RMS reconstruction error for shape or texture was calculated between the ground truth shape/texture and the reconstructed shape/texture of a missing training sample.
In this experiment, we randomly selected 30 subjects including 12600 face images to compare the reconstruction accuracy of M$^2$SA and CP-WOPT.
The incomplete shape and global texture tensors were obtained by randomly removing $10\%, 20\%, \cdots, 90\% $ and $95\%$ samples from the original complete shape and global texture tensors.
Then the proposed UT-AAM was created based on the completed shape and texture tensors using M$^2$SA and CP-WOPT and tested on the remaining 30 subjects with 12600 images.
We repeated this experiment 10 times and reported the average results.

Fig.~\ref{fig_7_1} and Fig.~\ref{fig_7_2} show the reconstruction RMS errors of the M$^2$SA and CP-WOPT methods using two different initialisation methods for completion of the incomplete shape and texture tensors, parametrised by the proportion of missing training samples. 
It is evident that the proposed initialisation method performs much better than the random initialisation method when using the M$^2$SA algorithm, in terms of both the average error and standard deviation of the mean.
In contrast, the CP-WOPT algorithm appears to be insensitive to different initialisation methods, until the proportion of missing samples is larger than $70\%$.
Note that the reconstruction error of the proposed initialisation method increases rapidly when the proportion of missing entries is higher than $80\%$.
The reason is that the restrictive nature of the `AND' operator results in entries being unavailable for initialising missing items and we switch to the `OR' operator.
However, the proposed initialisation method still performs better than that of the random initialisation method.
The CP-WOPT and M$^2$SA methods have similar performance when the proportion of missing entries is lower than $80\%$.
However, M$^2$SA performs much better than CP-WOPT when more than $80\%$ training samples are missing.
\begin{figure}[!t]
\centering
 \includegraphics[trim = 38mm 93mm 40mm 95mm, clip, width=.7\linewidth]{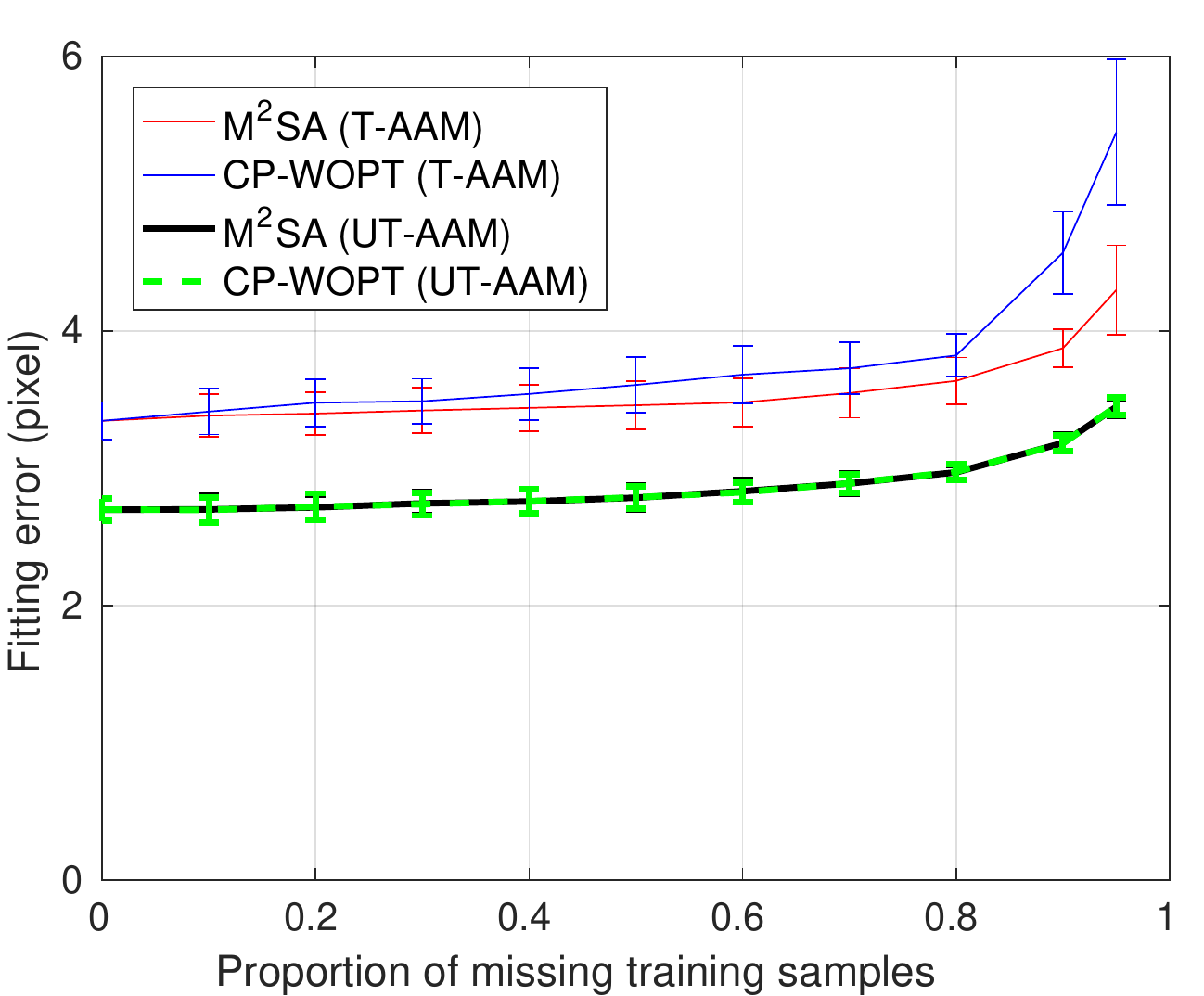}
\caption{A comparison of the M$^2$SA and CP-WOPT tensor completion algorithms in terms of the UT-AAM fitting error, parametrised by different proportions of missing training samples. We use our proposed initialisation method for tensor completion, and use the gradient-descent-based (GD) and proposed cascaded-regression-based (CR) algorithms for UT-AAM fitting.}
 \label{fig_7_3}
\end{figure}

We also evaluate the fitting error of UT-AAM parametrised by different proportions of missing training samples.
The results are shown in Fig.~\ref{fig_7_3}.
Note that, in this experiment, we also compared our UT-AAM with the classical T-AAM using the gradient-descent-based Gauss-Newton fitting algorithm.
First, it is evident that the proposed UT-AAM performs well even when a large proportion of training samples are missing.
Compared to the model constructed from a complete training dataset (the point `0' on the X-axis), the fitting errors grow slightly as the proportion of missing training samples increases.
Second, the M$^2$SA algorithm provides better fitting accuracy than CP-WOPT for the classical T-AAM.
In contrast, for the proposed UT-AAM using cascaded-regression-based fitting method, the difference in using M$^2$SA and CP-WOPT is minor.
The main reason is that the tensor completion algorithms are only used to build the tensor-based shape and texture model.
For cascaded regression based model fitting, the regressor was trained only from available training samples.
Last, an important finding here is that the proposed cascaded regression based fitting algorithm is more robust to the variation in the proportion of missing training samples.
As the proportion increases, the fitting error of the proposed UT-AAM using cascaded regression grows slowly.
In contrast, the classical T-AAM using the gradient-descent-based fitting algorithm is more sensitive to the proportion of missing training samples and has higher fitting error. 

\subsection{The use of UT-AAM in facial landmark detection}
\begin{figure}[t]
\centering
\subfloat[Original 7 pose variations]{
\label{fig_9_1}
 \includegraphics[trim = 0mm 105mm 0mm 0mm, clip, width=1\linewidth]{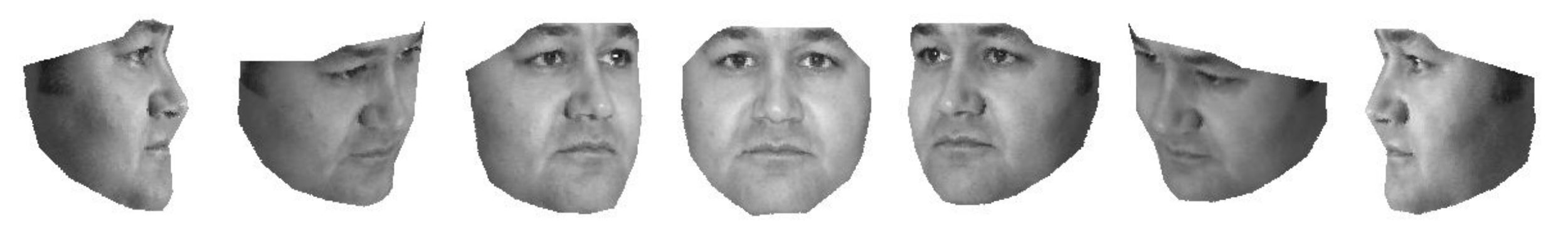}
}
\\
\subfloat[Synthesised additional 6 pose varitaions]{
\label{fig_9_2}
 \includegraphics[trim = 0mm 67mm 0mm 37mm, clip, width=1\linewidth]{./Img/fig9.pdf}
}
\caption{2D face instances synthesised using the proposed UT-AAM method: (a) the original 7 poses; (b) synthesised 6 additional poses.}
\label{fig_9}
\end{figure}

A potential use of our UT-AAM is through its capacity to generate 2D face instances.
In this section, we demonstrate the utility of UT-AAM for the training of 2D facial landmark detectors.
Recently, most cutting-edge facial landmark detection algorithms are data-driven and require a large number of training samples.
However, the laborious work of manually annotating facial landmarks for face images is tedious.
One alternative is to synthesise virtual training samples using a generative model, such as the 3D morphable face model~\cite{feng2015cascaded,kittler20163d,zhu2016face}.
However, the collection of 3D face scans and the construction of a 3D face model are very involved compared with the data collection and model construction of a 2D face model.
The most important advantage of UT-AAM is its capacity to generate realistic 2D face images by changing its mode-related model parameters. 
Fig.~\ref{fig_9} shows some examples synthesised by the proposed UT-AAM, by interpolating the tensor-based pose-mode coefficient vectors between two original faces with different poses.
We can see from the synthesised faces with new pose variations that UT-AAM is capable of performing realistic image synthesis.

Despite the capacity of UT-AAM to synthesise realistic face instances, the use of synthesised faces for facial landmark detector training presents some challenges.
As discussed in~\cite{feng2015cascaded}, synthesised faces are from different domains than real faces.
Synthesised face instances often lack complicated appearance variations in background and occlusion compared with real faces.
In that work, to gain a maximum benefit from synthesised faces, they proposed a Cascaded Collaborative Regression (CCR) that was trained on a mixture of real faces and synthesised faces by dynamically reducing the weights of synthesised training samples in the cascade~\cite{feng2015cascaded}.
Motivated by this, we compared the classical supervised descent method (SDM)~\cite{xiong2013supervised} with CCR for facial landmark detection.
For SDM training, we first used the training samples provided by the 300-W dataset, marked by `SDM (Real)'.
Then we used both the training images provided by 300-W and additional face instances synthesised by UT-AAM for SDM training, marked by `SDM (Real+Syn.)'.
Last, we trained the CCR model using both the real faces provided by 300-W and our synthesised faces (`CCR (Real+Syn.)').
In total, 46800 synthesised face instances were used, comprising 60 subjects with 3 expression, 20 illumination and $7+6$ pose variations (Fig.~\ref{fig_9}).
\begin{figure}
\centering
\subfloat[Outdoor, 68 points]{
 \includegraphics[trim = 10mm 80mm 20mm 85mm, clip, width=.7\linewidth]{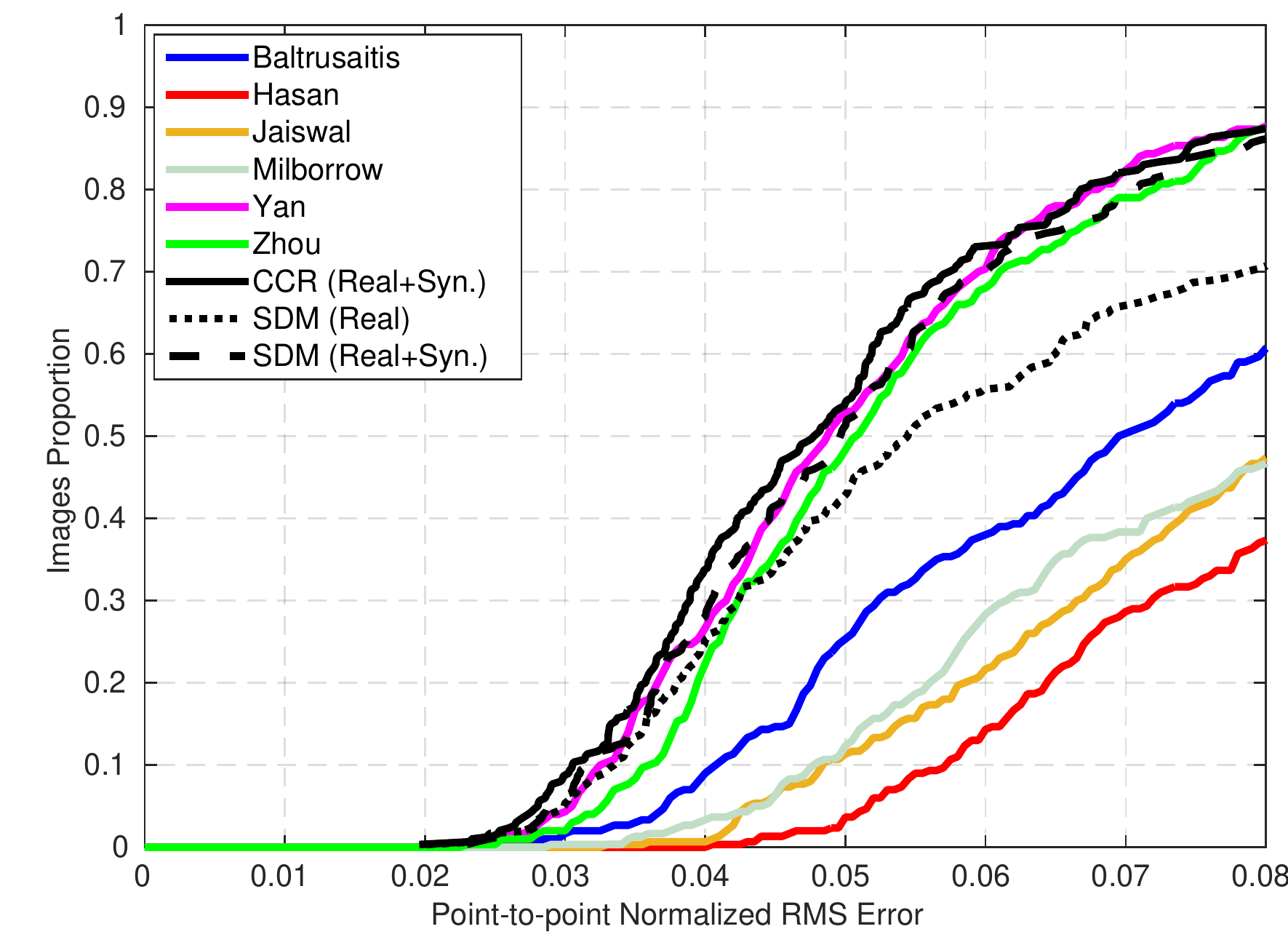}
}
\\
\subfloat[Indoor, 68 points]{
 \includegraphics[trim = 10mm 80mm 20mm 85mm, clip, width=.7\linewidth]{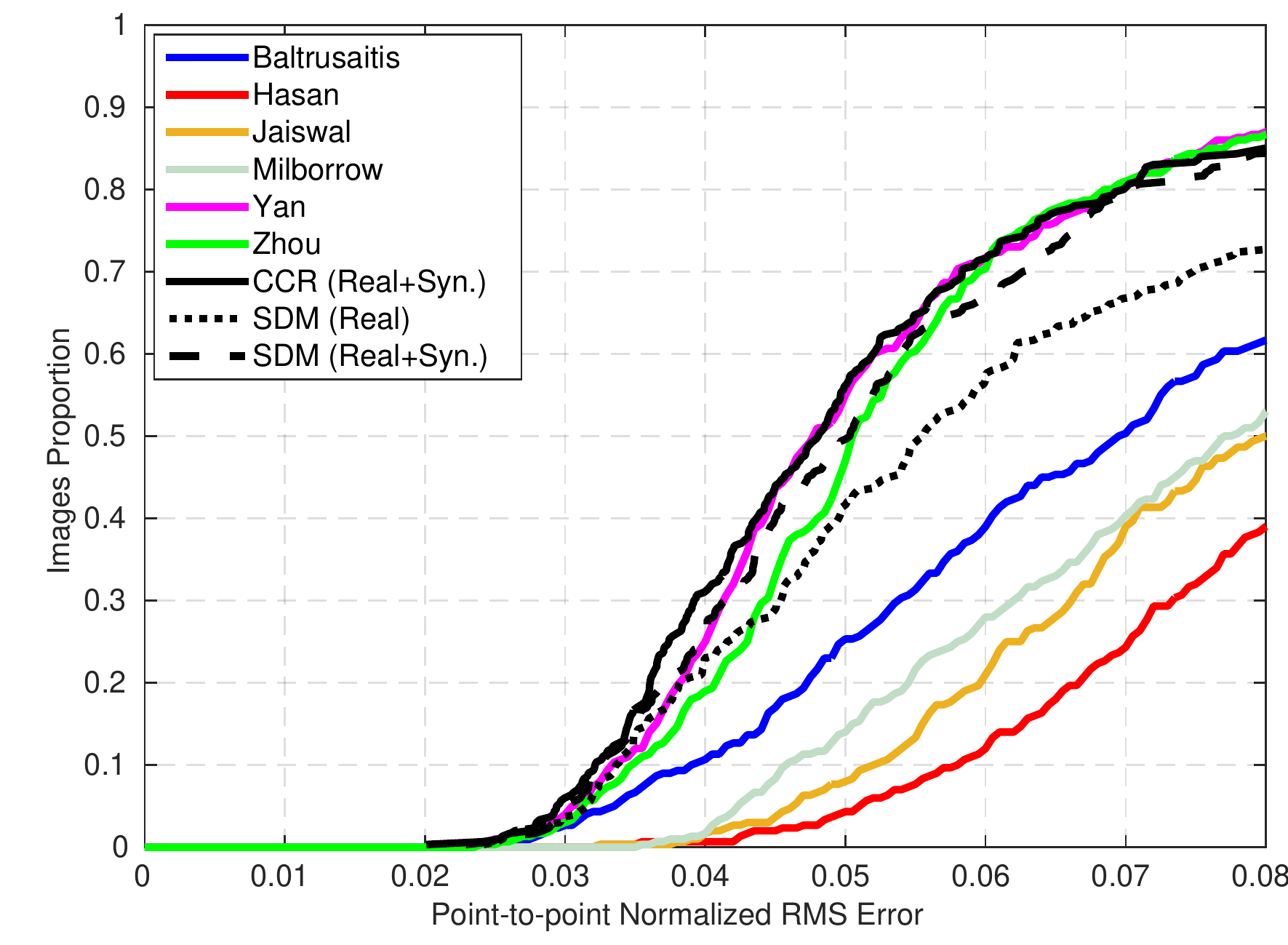}
}
\caption{A comparison of the cumulative error distribution curves of SDM and CCR, as well as a set of state-of-the-art methods from Baltrusaitis, Hasan, Jaiswal, Miborrow, Yan and Zhou~\cite{sagonas2016300}, on the 300-W face dataset: (a) results on the 300 outdoor face images; (b) results on the 300 indoor faces.}
\label{fig_10}
\end{figure}

The results obtained on the 300-W dataset are shown in Fig.~\ref{fig_10}.
It should be noted that, SDM is one of the most popular algorithms for detecting facial landmarks in unconstrained face images, and is usually used as a baseline.
CCR is an improved version of SDM, developed in particular for the purpose of using a mixture of real and synthesised faces.
Both CCR and SDM are based on a set of linear regressors in cascade.
Fig.~\ref{fig_10} shows that SDM performs well on the 300-W benchmark and beats most of the other algorithms.
In addition, the use of synthesised 2D face instances improves the performance of SDM significantly.
Last, the joint use of CCR and synthesised faces further improves the performance than SDM and beats all the other methods.

\section{Conclusion}
\label{Sec_Conclusion}
In this paper, we proposed a unified tensor-based AAM.
Compared with the classical tensor-based AAM, the proposed UT-AAM can be created from an incomplete training dataset and results in a unified single tensor model across different variation modes.
To deal with the problem of self-occlusion, a unified landmarking strategy was advocated for obtaining universal shape and texture representations of faces across large pose variations.
A more efficient and accurate cascaded-regression-based model fitting algorithm was also proposed for UT-AAM fitting.
Experiments conducted on the Multi-PIE face dataset demonstrate the merits of the proposed UT-AAM algorithm.
Last, we showed that the use of our UT-AAM to augment the volume of training data for a facial landmark detector training improved its performance on the 300-W benchmarking dataset.

More recently, powerful algorithms such as Deep Neural Networks (DNN) have been successfully used as weak regressors in cascaded regression, delivering impressive results in facial landmark detection~\cite{Trigeorgis_2016_CVPR}.
In future studies we plan to explore the merit of incorporating DNN and synthesised 2D faces in facial landmark detection.
One underlying assumption of the successful use of DNN is a big training dataset.
However, currently, no such a training dataset for facial landmark detection is publicly available.
We believe that the use of a large volume of synthesised training samples can also improve the performance of DNN-based facial landmark detection algorithms.

\section*{Acknowledgment}
This work was supported in part by the EPSRC Programme Grant `FACER2VM' (EP/N007743/1), the National Natural Science Foundation of China (61373055, 61672265) and the Natural Science Foundation of Jiangsu Province (BK20140419, BK20161135).

\ifCLASSOPTIONcaptionsoff
  \newpage
\fi



\bibliographystyle{IEEEtran}
\bibliography{fengref}
\end{document}